\documentclass[letterpaper, 10 pt, conference]{ieeeconf}  

\IEEEoverridecommandlockouts                              

\usepackage{graphics} 
\usepackage{graphicx}

\usepackage{epsfig} 
\usepackage{amsmath} 
\usepackage{amssymb}  
 \usepackage{algorithm}
\usepackage{algpseudocode}
\algnewcommand\AAND{\textbf{ and }}
\algnewcommand\Or{\textbf{ or }}
\usepackage{color}
\usepackage{citesort}
\usepackage{flushend}
\usepackage{url}

\DeclareMathAlphabet{\pazocal}{OMS}{zplm}{m}{n}

\usepackage{array}
\newcolumntype{C}[1]{>{\centering\arraybackslash}p{#1}}
\newcolumntype{M}[1]{>{\raggedright\arraybackslash}p{#1}}

\usepackage{array} 
\newcolumntype{L}[1]{>{\raggedright\let\newline\\\arraybackslash\hspace{0pt}}m{#1}}	
\newcolumntype{S}[1]{>{\centering\let\newline\\\arraybackslash\hspace{0pt}}m{#1}}
\newcolumntype{R}[1]{>{\raggedleft\let\newline\\\arraybackslash\hspace{0pt}}m{#1}}

\algnewcommand\pushup{\vspace{-1ex}}
\algnewcommand\pushuphalf{\vspace{-0.5ex}}

\makeatletter
\renewcommand*{\@opargbegintheorem}[3]{\trivlist
  \item[\hskip \labelsep{\itshape #1\ #2}] \textit{(#3)}\ }
\makeatother

\title{\LARGE \bf
Autonomous Aerial Robotic Surveying and Mapping with Application to Construction Operations
}

\author{Huan Nguyen, Frank Mascarich, Tung Dang, Kostas Alexis
\thanks{This material is based upon work supported by Governor's Office of Economic Development of the State of Nevada under the Construction Robotics Award.}
\thanks{The authors are with the Autonomous Robots Lab, University of Nevada, Reno, 1664 N. Virginia, 89557, Reno, NV, USA
        {\tt\small kalexis@unr.edu}}%
}

\begin{document}

\maketitle
\thispagestyle{empty}
\pagestyle{empty}

\begin{abstract}
In this paper we present an overview of the methods and systems that give rise to a flying robotic system capable of autonomous inspection, surveying, comprehensive multi-modal mapping and inventory tracking of construction sites with high degree of systematicity. The robotic system can operate assuming either no prior knowledge of the environment or by integrating a prior model of it. In the first case, autonomous exploration is provided which returns a high fidelity $3\textrm{D}$ map associated with color and thermal vision information. In the second case, the prior model of the structure can be used to provide optimized and repetitive coverage paths. The robot delivers its mapping result autonomously, while simultaneously being able to detect and localize objects of interest thus supporting inventory tracking tasks. The system has been field verified in a collection of environments and has been tested inside a construction project related to public housing. 
\end{abstract}

\section{INTRODUCTION}

Robotic systems, both on the ground and flying, are increasingly more frequently utilized in the construction industry~\cite{prieto2016universal,willmann2016robotic,helm2012mobile,saidi2016robotics,whittaker1986construction,ibrahim2017interactive,alejo2014collision,jimenez2013control,gheisari2016unmanned}. This paper outlines the early development of an integrated aerial robotic system capable of autonomous surveying and mapping tasks inside complex construction projects including the precise characterization of GPS-denied indoor structures. The solution integrates a set of contributions in the domains of a) multi-modal localization and mapping fusing LiDAR point clouds combined with visible-light and thermal cameras and inertial sensors, b) path planning both for autonomous exploration and optimized and repetitive inspection, alongside c) the ability to detect and localize objects of interest for possible inventory tracking tasks. Being a small flying platform, the system is unbound by terrain which is appealing in the often complex-to-traverse construction site environments.  

\begin{figure}[h!]
\centering
    \includegraphics[width=0.99\columnwidth]{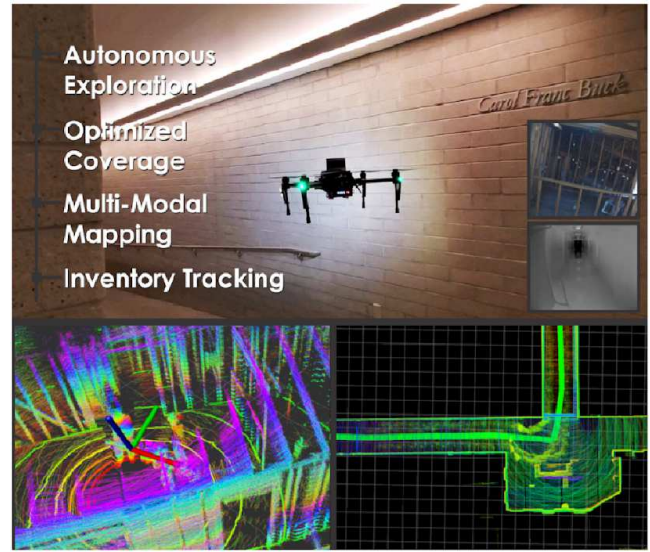}
\caption{Instances and indicative results of autonomous mapping nd characterization missions conducted using the Robotized Construction Site Surveyor system.}\label{fig:rcssintro}
\end{figure}

The designed Robotized Construction Site Surveyor (RCSS) - depicted in Figure~\ref{fig:rcssintro} - is the product of the integration and task-specific specialization of research conducted across domains and applications including those of nuclear site characterization~\cite{mascarich2018radiation}, subterranean exploration~\cite{GBPLANNER_IROS_2019}, industrial inspection~\cite{SIP_AURO_2015}, navigation in degraded visual environments including dust and smoke~\cite{khattak2019robust}, agile control~\cite{mpc_rosbookchapter} and more. The individual methods and systems developed for such tasks were adjusted and integrated for the development of this application-specific robotic platform. The resulting system has been tested and verified in a collection of field settings including complex building environments and a construction site relating to public housing. 

%
%

The remainder of this paper is organized as follows. Section~\ref{sec:perception} provides an overview of the designed multi-modal perception system responsible for localization and mapping, as well as object detection and localization. Section~\ref{sec:planning} outlines the two core path planning methods for autonomous exploration and optimized structural inspection respectively. Finally, Section~\ref{sec:exp} presents relevant experimental results, followed by conclusions in Section~\ref{sec:conclusions}. 

\section{MULTI-MODAL CHARACTERIZATION OF CONSTRUCTION SITES}\label{sec:perception}

A core component of the Robotized Construction Site Surveyor is that of multi-modal sensor fusion for localization and mapping, as well as object detection and position reporting. Deriving precise maps of construction sites is essential for building modeling and RCSS achieves this task through the fusion of LiDAR pointclouds, visible-light and thermal vision cameras, and inertial measurement cues. The motivation behind this multi-modal sensor fusion relates to two goals, namely to a) enable precise mapping even in the most degraded environments (e.g., during darkness or subject to significant dust and other obscurants), as well as b) provide a comprehensive mapping result delivering structural $3\textrm{D}$ information combined with vision-based texture data and thermal signature information. The latter, can be used to detect deficiencies in the construction or pre-existing problems in a building (e.g., in the case of a renovation project). Furthermore, these robot capabilities have been extended to include the ability to detect objects of interest and report their precise pose in the map so as to support inventory keeping and maintenance. 

\begin{figure}[h!]
\centering
    \includegraphics[width=0.99\columnwidth]{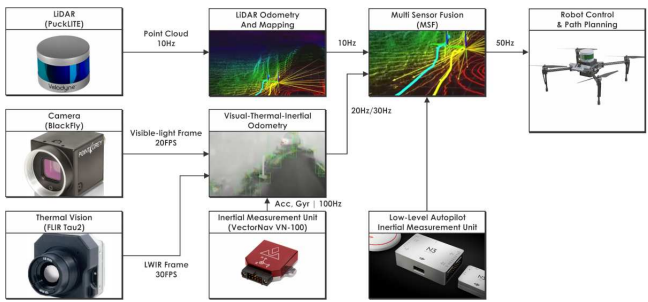}
\caption{Block diagram of the implemented multi-modal characterization and mapping solution of the Robotized Construction Site Surveyor. The methods have been developed in the framework of research first conducted for subterranean exploration.}\label{fig:multimodalmapping}
\end{figure}

The process of multi-modal localization and mapping is facilitated by a loosely coupled approach that combines (visible-light and longwave infrared) camera-inertial updates with LiDAR-based odometry and mapping as depicted in Figure~\ref{fig:multimodalmapping}. Camera-inertial updates are provided at high update rates (e.g., $20\textrm{FPS}$ or $30\textrm{FPS}$). Visible-light cameras provide accurate results in most environments with good illumination but suffer in conditions of darkness, lack of texture or presence of obscurants (such as dust which is common in construction sites). The Robust Visual-Inertial Odometry (ROVIO) method in~\cite{bloesch2015robust} is utilized for the purposes of camea-IMU based estimation. Thermal vision on the other hand penetrates most such conditions of visual degradation but in general is characterized of images of low contrast. The modified ROVIO method in~\cite{khattak2019robust,bloesch2015robust} is utilized for the purposes of thermal-inertial odometry estimation. When geometrically-rich structure is available in the environment, which is typically the case in construction projects, LiDAR Odometry And Mapping~\cite{zhang2014loam} can perform well in most conditions and only gets degraded when dealing with self-similar geometry, obscurants and other challenges. RCSS combines the individual pose updates in a loosely coupled fashion as outlined in~\cite{VIOLOAM_ICUAS2020} thus leading to a more resilient multi-modal approach.

%
%

\section{EXPLORATION AND STRUCTURAL INSPECTION PATH PLANNING}\label{sec:planning}

\begin{figure}[h!]
\centering
    \includegraphics[width=0.99\columnwidth]{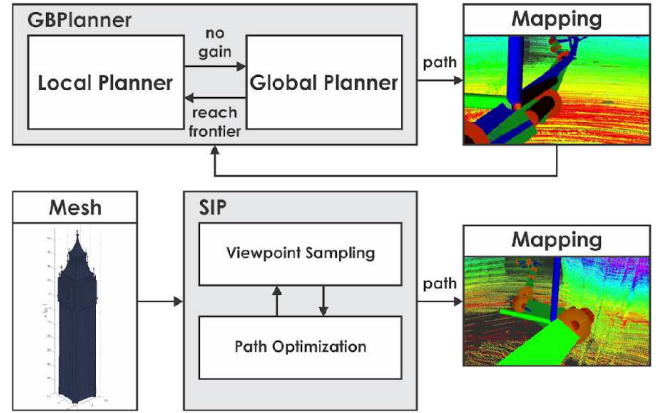}
\caption{Block diagram overview of the implemented planners for autonomous exploration and optimized structural inspection (given a prior model of the facility).}\label{fig:planners}
\end{figure}

Monitoring of construction projects relies on repetitive inspection and information processing. In particular, comprehensive $3\textrm{D}$ mapping, visual monitoring, coverage with task-specific sensing, and inventory handling are of special importance. To execute these tasks in an automated manner, RCSS implements two types of path planning, namely a) for autonomous exploration given no prior knowledge of the environment, as well as b) optimized coverage given a prior model of the structure of interest. An overview of the role of each planner is depicted in Figure~\ref{fig:planners}. 

%
%

For the first task, to be used when no prior model is available, the robot is employing the Graph-based Exploration Path Planner (GBPlanner) detailed in~\cite{GBPLANNER_IROS_2019}. The method uses a bifurcated architecture according to which it a) plans exploratory path by spanning a random graph in a local volume around the current robot and evaluating the volumetric gain of its of its vertices, while only when such an exploratory path cannot be identified it b) utilizes the incrementally built global graph to identify a previously detected frontier of the exploration space to which it moves to continue its mission. 

For the first task, to be used when a prior model is available either by blueprints/CAD or a previous exploratory mission, the Structural Inspection Planner (SIP) in~\cite{SIP_AURO_2015} is utilized. SIP is utilizing a mesh representation of the structure and then employs an iterative 2-step algorithm to find full coverage paths with minimum cost (time-of-travel), while respecting vehicle and sensor constraints. According to its 2-step paradigm it first samples a set of vertices that provide viewpoints to ensure full coverage of the structure by solving a $3\textrm{D}$ Art Gallery Problem (AGP)~\cite{lee1986computational}. It then finds an optimal full coverage path by solving the associated Traveling Salesman Problem (TSP)~\cite{junger1995traveling}. Notably, the edges between the AGP-sampled vertices are found using RRT$^\star$ so as to be collision free given the map of the environmnet. The whole process is iteratively repeated by trying to adjust the vertices such that lower cost and full coverage paths can be found. 

\section{EXPERIMENTAL VERIFICATION}\label{sec:exp}

To evaluate the performance and application potential of RCSS we conducted three field experiments. These in particular included a) the autonomous exploration and mapping within the corridors of the first floor of the Arts Building of the University of Nevada, Reno, b) the autonomous inventory reporting of objects of interest inside another building, as well as c) the mapping of a construction site related to a community housing project. The first two environments involve completed structures, while the second involves an environment with its underlying timber-based construction exposed and thus presents more complex geometry with thinner features. 

\begin{figure}[h!]
\centering
    \includegraphics[width=0.99\columnwidth]{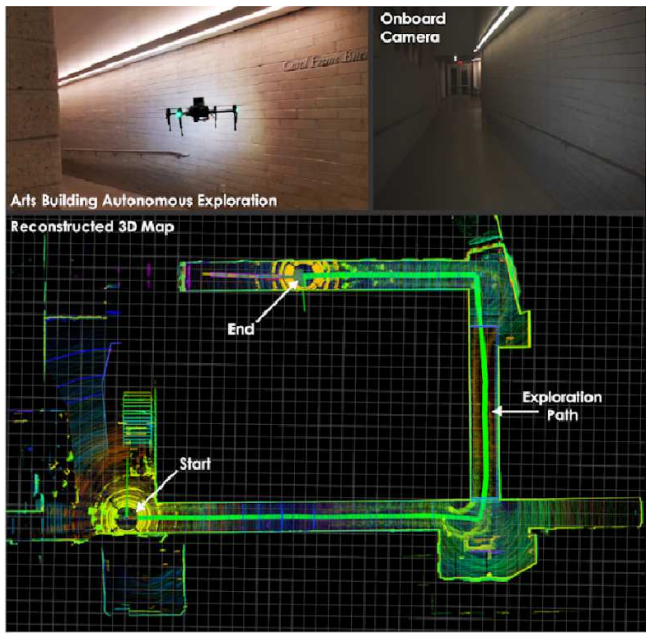}
\caption{Overview of the autonomous exploration and online $3\textrm{D}$ mapping mission inside the Arts Building of the University of Nevada, Reno. Such missions are needed when a new construction project is to start and updated structural models may be required.}\label{fig:artsbuilding}
\end{figure}

The results of the autonomous exploration within corridors of the Arts Building of the university are shown in Figure~\ref{fig:artsbuilding}. Such a mission is necessary when new activities in an existing building are to start and thus a map of the current geometric form has to be acquired. 

%
%

\begin{figure}[h!]
\centering
    \includegraphics[width=0.99\columnwidth]{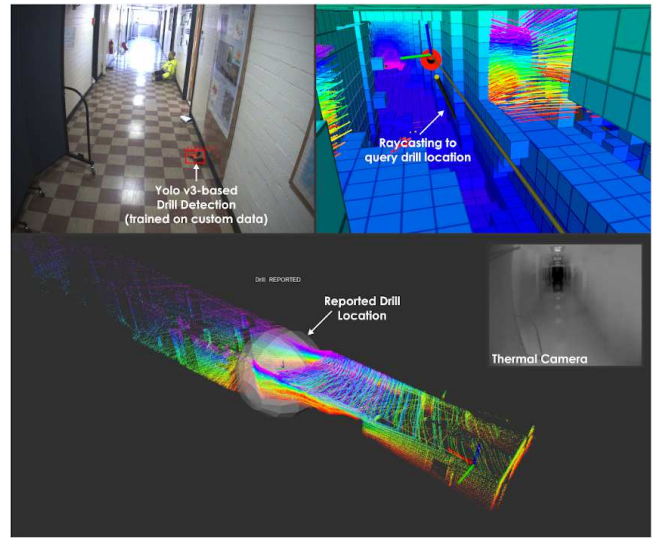}
\caption{Automated object of interest detection and localization in the online reconstructed $3\textrm{D}$ map for inventory management inside construction sites.}\label{fig:yoloinventoryreports}
\end{figure}

The next experimental study involved the use of the online mapping capabilities of the robot in combination with a Yolo v3-based trained object detector for objects of relevance to construction activities so as to perform automated inventory management. The robot navigates a building, detects the objects of interest and automatically reports their a) class, and b) location at c) a certain timestamp. Given the online reconstructed map of the environment, the object location is acquired through ray casting on the volumetric map representation. Details for this process can be found in~\cite{ARTIFACTS_AEROCONF2020}. An indicative result is shown in Figure~\ref{fig:yoloinventoryreports}. 

%
%

\begin{figure}[h!]
\centering
    \includegraphics[width=0.99\columnwidth]{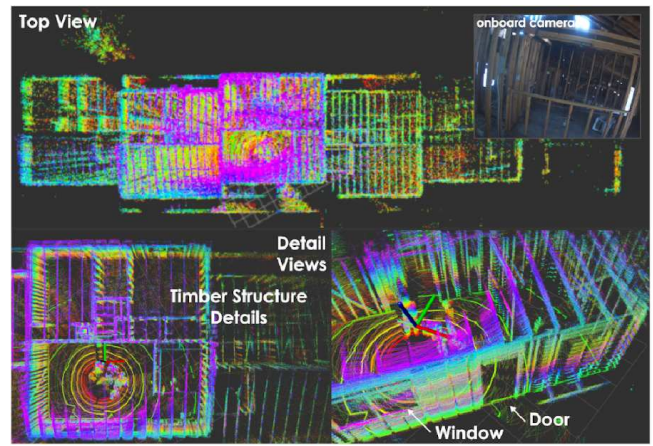}
\caption{Precise mapping of complex timber structure in the framework of a construction project related to community housing. The particular building was burned after a fire and the construction workers had removed all wall elements exposing the underlying timber structure.}\label{fig:communityhousingmap}
\end{figure}

We further verified the ability of RCSS to map in detail timber structures in the framework of an actual community housing construction project. The result is shown in Figure~\ref{fig:communityhousingmap} and indicates that precise metric maps of the structure can be acquired. This is particularly useful in such old buildings for which blueprints are not available. We conducted this mapping mission after construction activities had started. The particular building was burned due to a fire accident and the construction workers had removed all wall remainings thus exposing the underlying timber structure.

%
%

\begin{figure}[h!]
\centering
    \includegraphics[width=0.99\columnwidth]{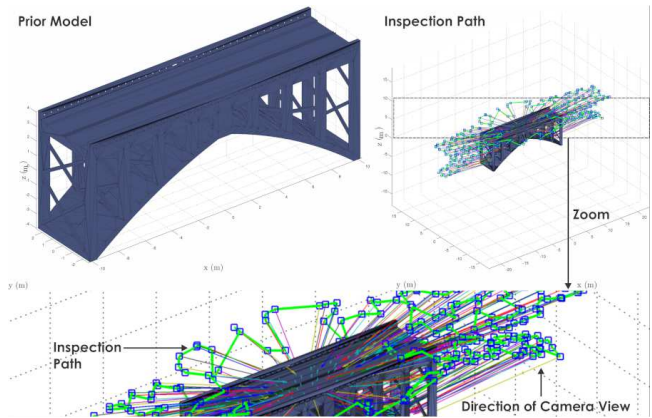}
\caption{Inspection path planning result ensuring full coverage given a prior model of a bridge. For full coverage planning, a camera sensor with constrained vertical and horizontal field of view is considered..}\label{fig:inspectpp}
\end{figure}

\begin{figure}[h!]
\centering
    \includegraphics[width=0.99\columnwidth]{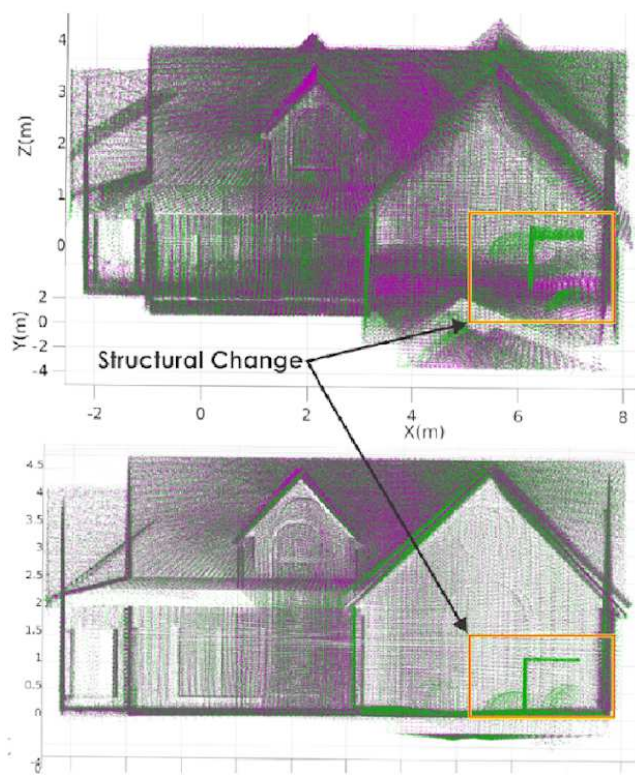}
\caption{Visualization of aligned point cloud based on two inspection missions on simulated data to allow observation of areas of change.}\label{fig:changedetect}
\end{figure}

Last we present two results based in simulated data. The first relates to the derivation of an inspection path given a prior model of a bridge structure as outlined in Section~\ref{sec:planning}. Figure~\ref{fig:inspectpp} presents the relevant result with a path ensuring full coverage and thus potential for systematic inspection. A modification to the open-source version of the planner mentioned in Section~\ref{sec:planning} was conducted to offer uniform coverage in terms of distance to each mesh facet. The second relates to performing alignment of point cloud data derived from different missions of environments in order to identify changes. The result - depicted in Figure~\ref{fig:changedetect} visualizes the two aligned point clouds and observation allows to detect what is different, a process which will be automated as RCSS progresses with its development. 

%
%

%
%

\section{CONCLUSIONS}\label{sec:conclusions}

This paper overviewed the current capabilities of the Robotized Construction Site Surveyor system. Although in its early phase of development, this aerial robot builds on top of extensive prior research in multi-modal localization and mapping and autonomous exploration and inspection path planning. Combining these robust and resilient algorithms and subsystems with dedicated object detection and location estimation for inventory tracking, as well as the ability to deliver multi-modal maps of construction facilities fully autonomously, rapidly and unbound from terrain limitations makes this system appealing for use in real-life. Ongoing and future research will emphasize on its utilization in active construction projects in the area of Northern Nevada. 

\bibliographystyle{IEEEtran}
\bibliography{./BIB/RCSS_BIB}

\end{document}